\documentclass[letterpaper, 10 pt, conference]{ieeeconf}  

\IEEEoverridecommandlockouts                              

\usepackage{graphics} 
\usepackage{epsfig} 
\usepackage{amsmath} 
\usepackage{amssymb}  
\usepackage{ upgreek }
\usepackage[linesnumbered,ruled]{algorithm2e}
\usepackage[utf8]{inputenc}

\usepackage{xcolor}

\usepackage{graphicx}
\usepackage{float}
\usepackage{amsmath}

\usepackage[official]{eurosym}

\usepackage{subfigure}
\usepackage{algpseudocode}
\usepackage{epstopdf}
\usepackage{multicol}
\usepackage[leftcaption]{sidecap}
\usepackage{booktabs}

\usepackage{amssymb}
\usepackage{arydshln}

\usepackage[hidelinks]{hyperref}
\usepackage{paralist}

\title{\LARGE \bf
Underwater Soft Robotic Hand  with Multi-Source Coupling Bio-Inspired Soft Palm and Six Fingers Driven by Water Hydraulic
}

\author{Haihang Wang, He Xu$^*$, Chen Yang, Xin Li, and Siqing Chen
\thanks{The authors are  with College of Mechanical and Electrical Engineering, Harbin Engineering University, China. Corresponding to He Xu: e-mail: railway\_dragon@sohu.com}
}

\begin{document}

	\maketitle
	\thispagestyle{plain}
	\pagestyle{plain}

	\begin{abstract}
	A new fluid-driven soft robot hand in this study uses the idea of the bionics and has the anthropomorphic form, which is oriented to the flexible grasp function. The soft robot hand is composed of a new kind of multi-freedom soft finger and soft palm, which realizes the characteristic grasping function of forehand and backhand. Combined with the fine fluid control system, the soft hand can realize flexible grasping under high pressure, so as to realize flexible grasping operation for different types of target objects in the underwater environment. The soft robot hand was controlled based on water hydraulic platform, Finally, the soft robot hand and the fine fluid control system were connected to form the underwater soft robot hand experiment platform.
	
	\end{abstract}

\section{Introduction}
Robotics research is becoming increasingly involved in different directions of scientific research, new robotic platforms are being developed. Different forms of innovation and challenges are emerging, and new scientific discoveries are becoming possible. More and more technological inspiration of human innovation comes from natural features. Bionic robot refers to the transformation of biological principles into engineering design rules, which have different challenges that must be solved, and also make innovation more in line with nature and life \cite{RN1743}. Human-Robot Interaction (HRI) has been a topic of wide concern in academic circles. The development of robots is the cornerstone, autonomous movement and information exchange are the development \cite{RN1745}.

Traditional rigid robots have been widely used in manufacturing. Because their overall structure is rigid, they have disadvantages such as a single task and limited adaptability, and they are unsafe to interact with humans. At present, soft robot has gradually attracted more and more scholars and laboratories to carry out research and breakthrough. The function and information exchange capability of soft robot cannot be replaced by traditional rigid robot. In contrast to rigid robots, the soft robot is made of a soft and stretchable material, such as the silicone rubber. The soft robot has a continuous deformable drive structure similar to muscle, which can simulate the biological system to achieve different high curvature bending and twisting, and achieve relatively large degree of freedom. This makes the soft robot show unprecedented adaptability and agility. Compared with rigid robots, soft robots can operate in a limited space. Because of its structure and materials, soft robots are highly adaptive and can safely interact with humans and the natural environment. The human-computer interaction capability of soft robots is also an increasingly developing research and application field, which includes many challenging problems. Recent advances in the field of soft robotics have made it possible to accelerate the development of digital design and manufacturing tools with the potential for solutions that have a positive social impact \cite{RN1746}.
\begin{figure}[tbp]
    \centering
    \includegraphics[width=1.0\linewidth]{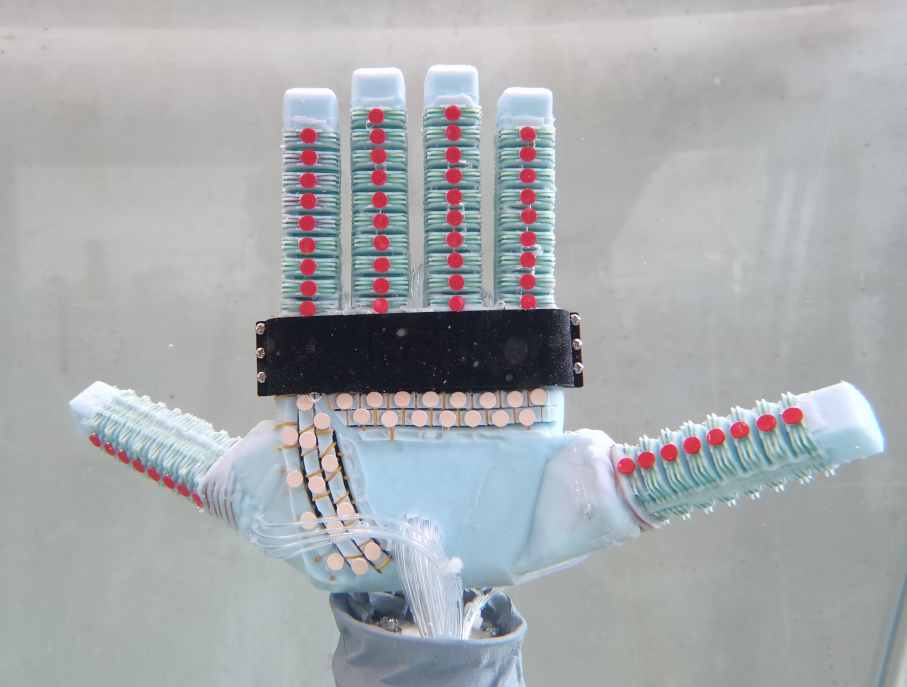}
    \caption{The underwater soft robotic hand in this study.}
    \label{fig:SHprototype}
\end{figure}

\begin{figure*}[ht]
    \centering
    \includegraphics[width=1.0\linewidth]{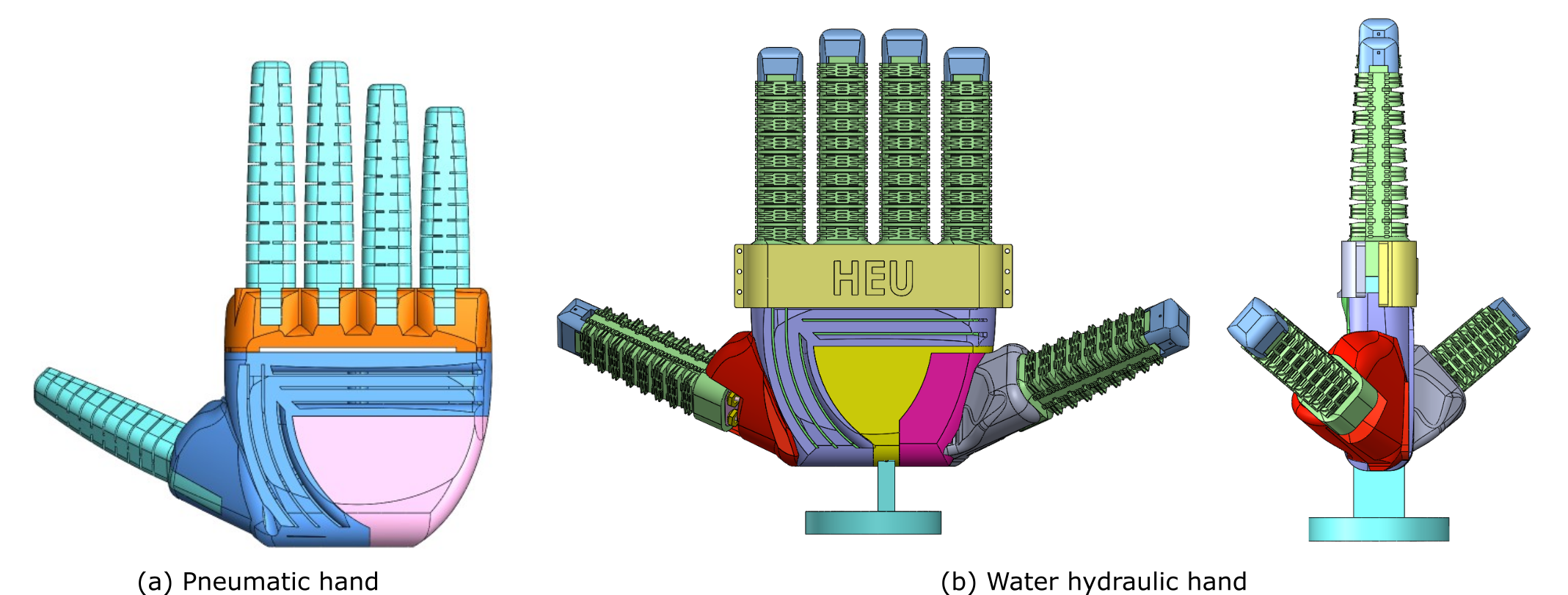}
    \caption{The Design of the pneumatic hand in Ref. \cite{RN1759} and the water hydraulic hand in this paper.}
    \label{fig:SHOverview}
\end{figure*}

The material and actuator of soft robot are the key points of its composition and structure, which directly affect its driving speed and bearing pressure. At present, the popular soft robot drive technology is the use of compressed air and pressurized fluid mode. This technology provides a powerful drive system for soft materials. Compressed air and fluid can directly inflate the chamber by using the channel network in the elastomer. The contraction device of McKibben actuator can generate relatively high force and displacement after deforming through compressed air, and make the robot move accordingly in combination with other structures of the robot \cite{wallace2008}. Worms move by using cylindrical parts of their bodies to produce traveling waves that contract and expand. The worm robot uses Pascal's principle to convert forces and displacements to achieve mechanical movements. By contracting the longitudinal muscles, they shorten the body and thus increase its diameter, and by contracting the circumferential muscles they lengthen the body by decreasing the diameter. Many worm-like robots are developed based on hydrostatic structures, and worm-like robot crawlers use shape memory alloy (SMA) actuators, most of which are soft-hard combined structures \cite{Quillin661}. Some complex soft-bodied movements are very attractive models, such as octopuses and squid. Flexible arms based on the anatomy and mechanism of octopus arm movement, highly deformable mechanical structures are woven from plastic fibers. The soft actuators are arranged laterally and longitudinally, and the overall bending is achieved by the longitudinal actuators. The flexible arm can shorten, elongate, bend or twist. Mimicking the actual state of an octopus, the muscle tissue maintains a constant volume, using hydrostatics to exchange displacement and force. The flexible arm works in water, taking advantage of the interaction with the environment \cite{Margheri2012,Mazzolai2012}. Other solutions for soft manipulators, such as the Octarm robot, use pneumatic muscles to bend in all directions \cite{RN1760}. The internal elastomer contains embedded channels that can be pressurized by fluid or air. Through careful design of chamber size, wall thickness and geometry, the charging and pressure relief of these internal channels can result in a variety of shapes of the software manipulator \cite{Shepherd20400}. Soft robots can treat the entire structure as a continuum. In order to minimize the force required to cause deformation, low modulus materials shall be used. Silica gel is a very good material, it not only has the characteristics of low modulus, but also has the characteristics of high strain and convenient production. The whole structure of the soft robot is continuous, and the deformation is highly nonlinear due to the large strain. The field of soft robotics is also creating computing and sensing materials that allow soft robots to operate better. Different environments also require different software selectors. Accordingly, soft robots can also follow the principles of artificial intelligence to enable us to develop biologically inspired artificial intelligence, which rigid robots cannot do \cite{RN1747,RN1748,RN1749}.

\section{Related Work}
\begin{table*}[htbp]
  \centering
  \caption{Technical parameters of the silicone materials}
    \begin{tabular}{cccccccccc}
    \toprule
    \textbf{Silicone} & \textbf{Shore Hardness} & \multicolumn{4}{c}{\textbf{Tensile Strength}} & \multicolumn{2}{c}{\textbf{100\% Modulus}} & \textbf{Elongation at Break (\%)} & \textbf{Cure type} \\
    \midrule
          &       & \textbf{psi} & \multicolumn{2}{c}{\textbf{N/mm2}} & \textbf{MPa} & \textbf{psi} & \textbf{kPa} &       &  \\
    \midrule
    Dragon Skin 10 & 10A   & 475   & \multicolumn{2}{c}{3.28} & 3.28  & 22    & 151   & 1000\% & Platinum \\
    Dragon Skin 30 & 30A   & 500   & \multicolumn{2}{c}{3.45} & 3.45  & 86    & 593   & 364\% & Platinum \\
    Smooth Sil 936  & 36A   & 550   & \multicolumn{2}{c}{3.8} & 3.8   & 180   & 1241  & 500\% & Platinum \\
    Elastosil M 4601 & 28A   & 942   & \multicolumn{2}{c}{6.5} & 6.5   & 54-65 & 375-450 & 700\% & Platinum \\
    VytaFlex 20A & 20A   & 200   & \multicolumn{2}{c}{1.38} & 1.38  & 50    & 345   & 1000\% & Urethane \\
    EcoFlex 00-30  & 00-30 & 200   & \multicolumn{2}{c}{1.38} & 1.38  & 10    & 69    & 900\% & Platinum \\
    Mold Star 30 & 30A   & 420   & \multicolumn{2}{c}{2.9} & 2.9   & 96    & 662   & 339\% & Platinum \\
    \bottomrule
    \end{tabular}%
  \label{tab:Silicone}%
\end{table*}%

The core function of an effective soft robot design is grasping. Because the soft robot with continuous structure breaks the structure of the connecting rod joint, the shape of the soft robot hand is more universal than that of the rigid robot hand. Combined with its own flexible materials to achieve a new more secure and effective grasping strategy and interactive functions.  Artificial muscle hydrostatic device design and intelligent muscle alignment of octopus-like robotic arm, can achieve bending, shortening and elongation functions. Through the combination of different longitudinal actuators, global bending can be obtained in each direction, and then grasping function can be realized \cite{RN1751}. At the same time, the internal cables will lead to the shortening and recovery of the robotic arm. The relationship between elongation and diameter reduction corresponds to the model of braids. A universal gripper composed of granular materials wrapped in elastic film can quickly grab and release various objects by using a combination of positive and negative pressures \cite{RN1752}. You can grab objects that are challenging for general purpose grippers, such as flat objects, soft objects, or objects with complex geometries. The gripper passively conforms to the shape of the target object, grabs the object in a vacuum hardening rigid way, and then uses the positive pressure to reverse back to the original form to achieve the effect of releasing the object. Human hands also offer great inspiration for adaptive robots. The joints maintain passive finger compliance, giving the fingers flexibility and an extended range of motion. Passive flexibility is the result of the direction of motion determined by the stiffness and geometric constraints of the joint. Using anatomically correct phalanx, elastic ligaments, and antagonistic tendons, an anthropogenic joint with multidirectional passive compliance was constructed. Fingers are directly used in coordination, which is conducive to mechanical adaptation when grasping large objects \cite{RN1750}. Professor George M. Whitesides Group developed a soft robot, which is driven by a pneumatic net formed by small channels, can achieve simple operation and complex movement. The whole structure design is made of elastomer material, which can realize the fast and large movement of the pneumatic network, and has a lower fatigue and failure rate. In order to achieve the multi-angle bending effect, a non-retractable material is needed to suppress the expansion of the cavity wall. The combination of high actuation rate and high reliability of soft robot actuator opens up a new application field for its expansion rate, channel shape and structural performance, thus achieving the advantages of fast inflation, easy control and measurement, light weight and so on \cite{RN1754}.

Soft robot hand is made of soft material, which has the advantages of strong flexibility, good man-machine interaction and simple structure and easy control. As an important branch of soft robot, soft manipulator has been gradually studied by more and more researchers and applied to people's production and life. The ability of soft robot hand to bend, elongate, contract, expand and twist is also an important aspect of research work at home and abroad. Wearable robotic hands have been used in medicine to drive the motor function of disabled hands and carry out strength training. The integral exoskeleton glove is a soft exoskeleton with an underactuated tendon actuation mechanism in the adaptive mechanism. Algorithms are used to limit the maximum level of muscle activation to prevent muscle and joint damage. Impedance-controlled rehabilitation exercises can provide different stimulation to muscles through different combinations of impedance parameters and adjust the exercise intensity by changing the control gain of the motion program controller \cite{RN1753}. Min Li team from Xi'an Jiaotong University \cite{RN1757} designed a robot teleoperation soft hand based on tactile feedback of fingertips is used to complete complex tasks and ensure the safety and friendliness of human-computer interaction. Manipulator can achieve finger flexion and contraction action. Bend sensors are embedded in the finger to monitor the bending Angle of the actuator. A pressure sensor on the fingertip feeds contact forces back to the operator. The manipulator is used for grasping fragile items, rescuing the injured, remote interaction with patients and other complex work, and the precision of the movement is very high. Manipulator demonstrates the application potential of flexible robot hands in complex operation tasks. Fluid driven flexible manipulator is highly compliant and can adapt to various objects. The tendon-driven flexible manipulator has the advantages of simple control, high flexibility, large output, but poor compliance. The fluid-driven manipulator improves this shortcoming. Most manipulator designs have only one drive method. A simple two-mode actuator is proposed that combines fluid and tendon actuation \cite{RN1756}. These brakes play to each other's strengths and compensate for each other's limitations, but design and control become more complex. By twisting the tube to drive, only one motor is needed to produce two drive modes. The finite element part of the actuator ensures the flexibility and softness of the fingers during grasping. The bending Angle, bending force and grasping ability of the flexible manipulator are improved, but the disadvantage is that the volume is large and the weight is large, which is not conducive to the wearable of soft robot. Deimel et al. of the Technical University of Berlin proposed a highly supple, underdriven, robust and dexterous anthropomorphic hand. This software hand has the advantage of being cheap to manufacture and adaptable to form. Efficient hand design with PneumFlex actuator. The effective dimension of grasping posture exceeds the dimension of driving signal, which proves that relatively simple control can achieve complex grasping behavior \cite{RN1755}. Yang Yang's team at the University of Hong Kong has designed a novel design of soft robot hand. The robot is 3D-printed and consists of five fingers and a palm. All fingers can be driven independently, and sensing modules are built into the entire hand structure. The fingers are driven by compressed air. In the bending process, the chamber is filled with loosely arranged particles and squeezed by air, which makes the fingers have high stiffness and increase the output force and carrying capacity \cite{RN1758}.

\section{Design}
\subsection{Soft Robotic Hand}

Our previous work in Ref. \cite{RN1759} focused on the design of a novel soft robotic hand with an biomimetic morphology. Its soft palm is designed in modular design approach, which includes two parts: Part A and B. The inspiring function of applying the soft actuator as the palm of the soft hand was presented and tested. The soft palm with inspiring function by applying the soft actuator in human hand-like motion performed a diversity of grasps.

The first generation and the second generation of the soft robot hand design are shown in Figure \ref{fig:SHOverview}(a). The relevant research of the first generation of the pneumatic soft hand \cite{RN1754} provides the preliminary configuration design and experience. It is pneumatic and the driving pressure range is generally between 0-50 kPa, the max pressure value in it can reach 70 kPa under the ultimate load. However, too high pressure will lead to the rupture risk of soft airbag of the hand. The low actuation pressure limits the soft hand's ability to grasp heavy objects.

Due to the shortcomings of the pneumatic soft hand, a soft bionic hand was designed which is driven by the water pressure. The new type soft hand can bear 300 kPa water pressure in the underwater environment, which has the large grasping force output for the underwater target. Figure Figure \ref{fig:SHOverview}(b) is the water hydraulic soft hand in the study, which is driven by water hydraulic pressure and works in an underwater environment. The main performance requirement of soft robot actuator is to inflate. The soft robot hand can meet the performance requirements of inflating, bending and twisting based on aerodynamic deformation.

\subsection{Material}

Silicone material is the key factor to determine the pressure resistance of the soft actuator. Tab. \ref{tab:Silicone} lists the commonly used silicone materials and the key technical parameters in the production of soft robots in the literature. Ecoflex 00-30 has a minimum shore hardness of 00 at 30, which means a very low hardness, equivalent to a racket or gel insoles. With a minimum tensile strength of 200 psi and a minimum 100\% modulus of 10 psi, Ecoflex 00-30 is the most stretchable and flexible material. The Dragon Skin 10 has a tensile strength of 475 psi and a maximum elongation at break of 1000\%, which reflects better toughness and high flexibility. Based on the experience of the first and second generation pneumatic soft hands, the S11 finger made with the Platinum-cured silicone of the Dragon Skin 10 has a 270 degree bend Angle output at 50 kPa. The shore hardness of the material is the core parameter that determines the pressure required to achieve the same bending angle.

The reason why the soft actuators break is not only that the water pressure is too high, but that the air cavity in the hand expands indefinitely without constraint, which makes some structural weak points continue to be stretched and then break. In order to improve its pressure bearing capacity, the third generation of soft hand was made of Smooth-Sil 936 two-component platinum cured silicone produced by Smooth-On Company, with a shore hardness of 36A, a natural color of blue, and a tensile strength of 550 psi, which was the ideal silicone material for this study.

\subsection{Sealing method}

Fluid-driven soft robots, when driven by the low pressure, basically use silicone self-sealing type that is that the trachea is directly inserted into the air cavity and the sealing effect is achieved by relying on the elastic deformation of silicone itself. Under high pressure, relying on the elasticity of silicone itself cannot be sealed, so the rigid joint is needed. The soft actuators in this paper are all sealed with the inverted taper joint with the screw fastening. The sealing structure of the reverse cone joint is composed of the reverse cone plug, the reverse cone ring and the matching Teflon hard tube. When in use, the end face of the hard tube is cut flat, and the reverse cone plug and the reverse cone ring are set in turn, and the bottom of the pressing ring is flush with the end face of the hard tube. When the reverse cone plunger is tightened, the reverse cone plunger compresses the reverse cone ring so that the ring closes the joint. The backward taper plug thread is 1/4 - 28UNF thread, and the sealing strength is 0.7 MPa.

\subsection{Soft Finger}
\begin{figure}[htbp]
    \centering
    \includegraphics[width=1.0\linewidth]{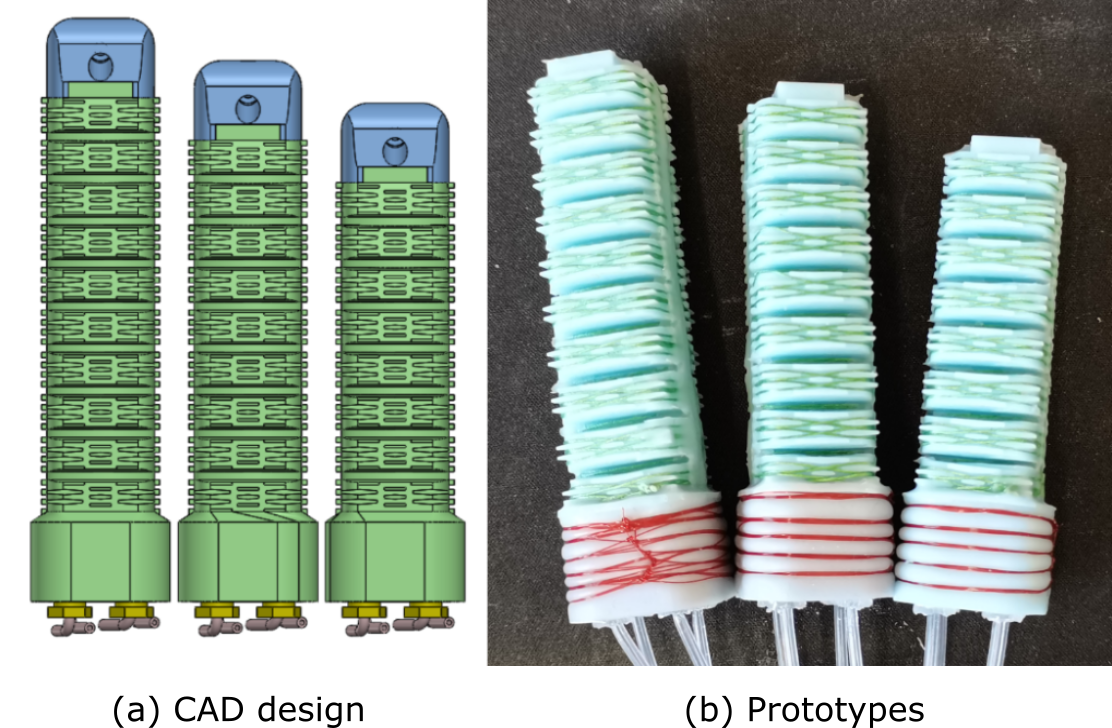}
    \caption{The CAD model and prototypes of the soft fingers}
    \label{fig:SFDesign}
\end{figure}

The soft fingers of this work adopt the bionic appearance design, and the overall configuration refers to the shape and size of human fingers, which is shown in Figure \ref{fig:SFDesign}. Its driving structure refers to the advantages of the Pneunets actuator and the fiber enhanced actuator. The fiber reinforced structure is added to the grid-shaped finger main structure in order to restrain the radial invalid deformation and the soft actuator can maximize its effective deformation potential.

\begin{figure*}[htbp]
    \centering
    \includegraphics[width=0.7\linewidth]{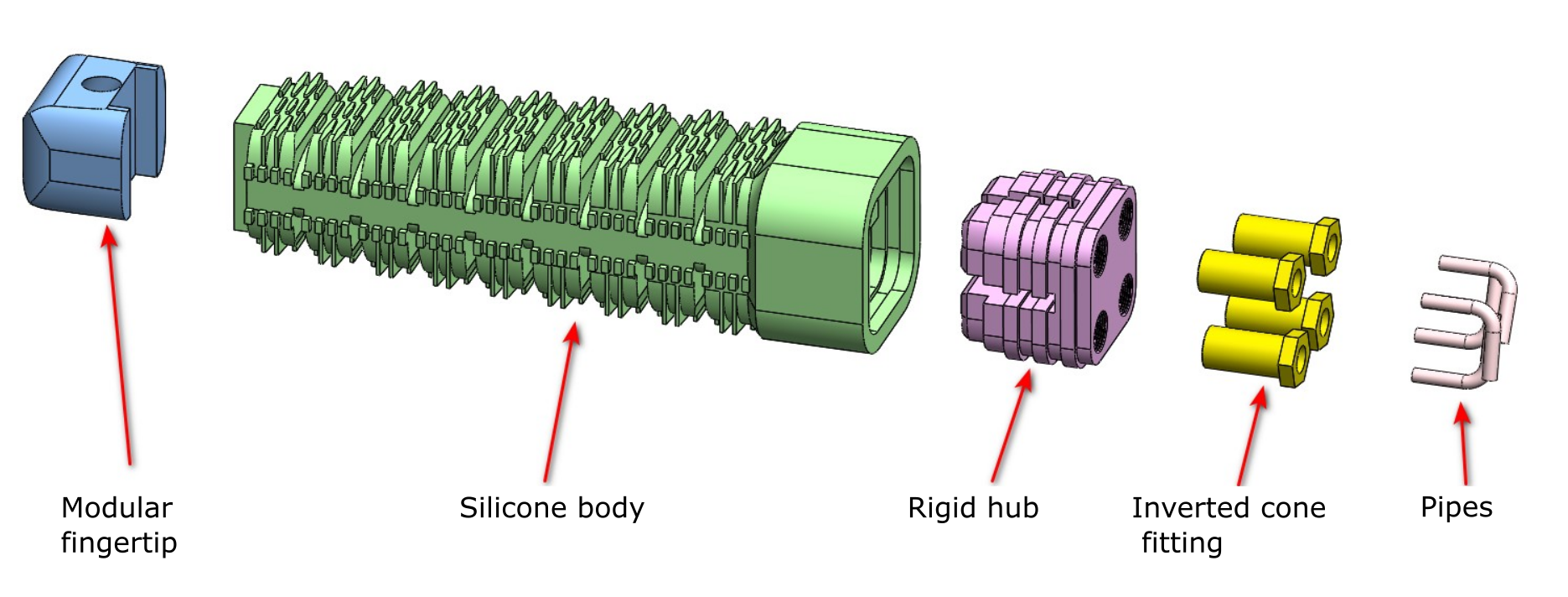}
    \caption{The Structure of the soft finger}
    \label{fig:SFstructure}
\end{figure*}

Figure \ref{fig:SFstructure} shows the structural design of the soft finger, which consists of an expandable and modular fingertip, a soft finger body, a sealing plug, an inverted tapered joint combination structure and a hard tube. The soft finger body is composed of a Smooth Sil 936 silicone. The sealing plug is based on the 3D printing technology and is made of the imported nylon material. The 1/4-28UNF tap is used to tap the four holes reserved to match the reverse taper joint seal structure. Taper joint components are produced and sold by Nanjing Runze Fluid Company, China.
The soft finger contains four symmetrical independent software cavities, which requires four sets of inverted cone seal combinations. Four groups of inverted cone joints with inverted cone combination realize the independent drive of four independent cavities. The gap between the sealing plug and the body of the soft finger is sealed by tightening and winding the thin wire, which is the red rope in Figure \ref{fig:SFstructure}(b).

\begin{figure}[htbp]
    \centering
    \includegraphics[width=0.9\linewidth]{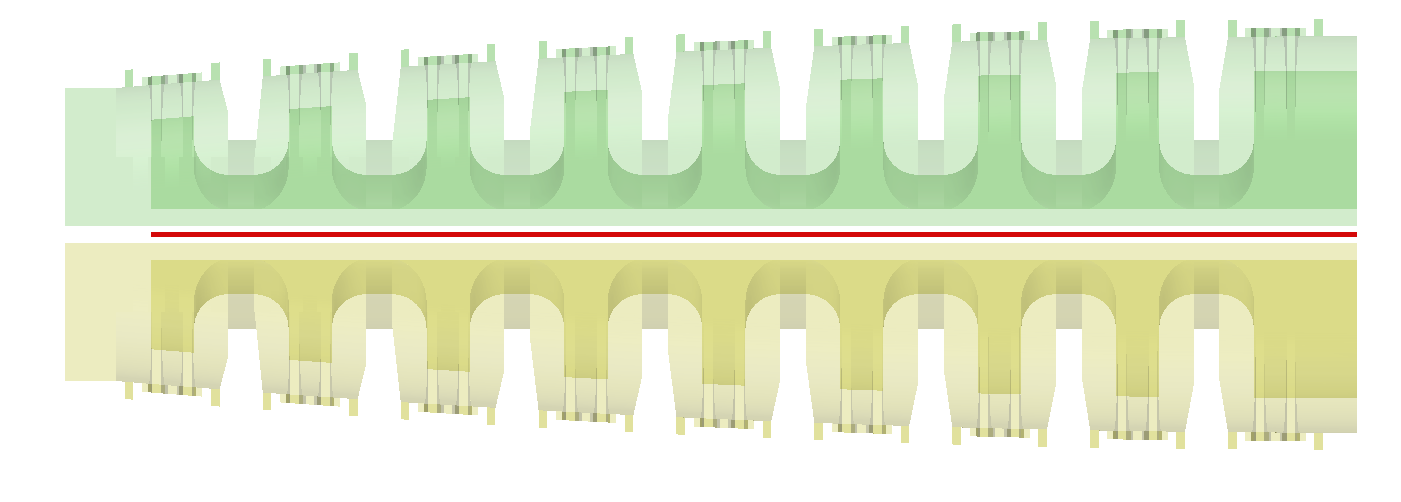}
    \caption{The inner chambers of the soft finger}
    \label{fig:SFinner}
\end{figure}

The fiber reinforced structure is the key structure to improve the pressure resistance of the soft finger. It restricts the expansion and deformation of the soft air cavity in the radial direction through regular cross winding, so as to maximize the expansion and bending ability of the soft air cavity in the axial direction. In Figure \ref{fig:SFstructure}(b), the green thin line is the fiber rope which is a super tensile subline. It is non-extensible, imported raw wire rope of 12 brails and can withstand a breaking ultimate tension of 164 N.

Figure \ref{fig:SFinner} shows the auxiliary groove structure on the body of the soft finger to guide the winding path.
The main body of the soft finger silicone is a four-cavity symmetrical design with a wall thickness of 2mm, as shown in Figure 10. The upper part of the green and the lower part of the yellow are the symmetrical structures. The green is an inextensible limiting layer and the material is organza.

\subsection{Soft Palm}
For the hand grasping, the palm always works together with the fingers to realize various operations. This paper discusses the function of soft palm from the perspective of human palm movement. Two main functions can be achieved, which are the thumb abduction and the palm bending. An important feature of the human hand is the abduction and adduction of the thumb, which allows the thumb to rotate and adjust position and orientation relative to the object being manipulated, it is essential for grasping.

Bending the entire plane of the palm helps to grasp smaller objects. The thumb of a soft hand cannot touch the other fingers without the aid of palm bending, even though all fingers are bent 180 degrees under the highest pressure. The software palm adopts modular design method. As shown in Figure 11, the structural design of the palm was inspired by the mechanism of the pneumatic mesh actuator which is consisted of a mesh chamber inside a silicone resin on a strain-limiting layer. It realizes the abduction of the thumb and the overall bending of the palm. The bending motion of the palm is achieved by enlarging the horizontal width of the pneumatic net actuator. The width of the air chamber along the palm is designed to be as wide as possible to generate enough force to bend the palm along the distal fingers. In addition, the humanoid aesthetic design is realized in the last step after ensuring that the soft actuator works as expected and does not affect the performance of the pneumatic actuator.

\section{Experiment}
\subsection{Test of the soft finger}

\begin{figure*}[htbp]
    \centering
    \includegraphics[width=0.85\linewidth]{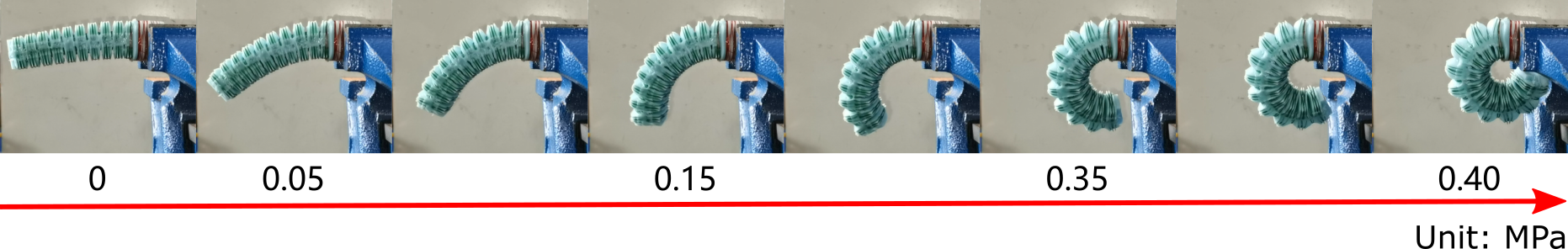}
    \caption{The test results of the soft finger when actuating single chamber}
    \label{fig:SFtest}
\end{figure*}

The soft finger used in this study was tested to analyze its performance on bending and grasping. The bending motions are shown in Figure \ref{fig:SFtest}. One of the four individual chambers was driven from 0 to 0.40 MPa. The finger can bend up to 270 degree when the pressure was 0.40 MPa. In general, the soft fingers are designed for being actuated under 0-0.35 MPa pressure. The finger under 0.35 MPa can bend 180 degree, which meet most situation in real world.

\subsection{Test of the soft palm}
\begin{figure}[htbp]
    \centering
    \includegraphics[width=1\linewidth]{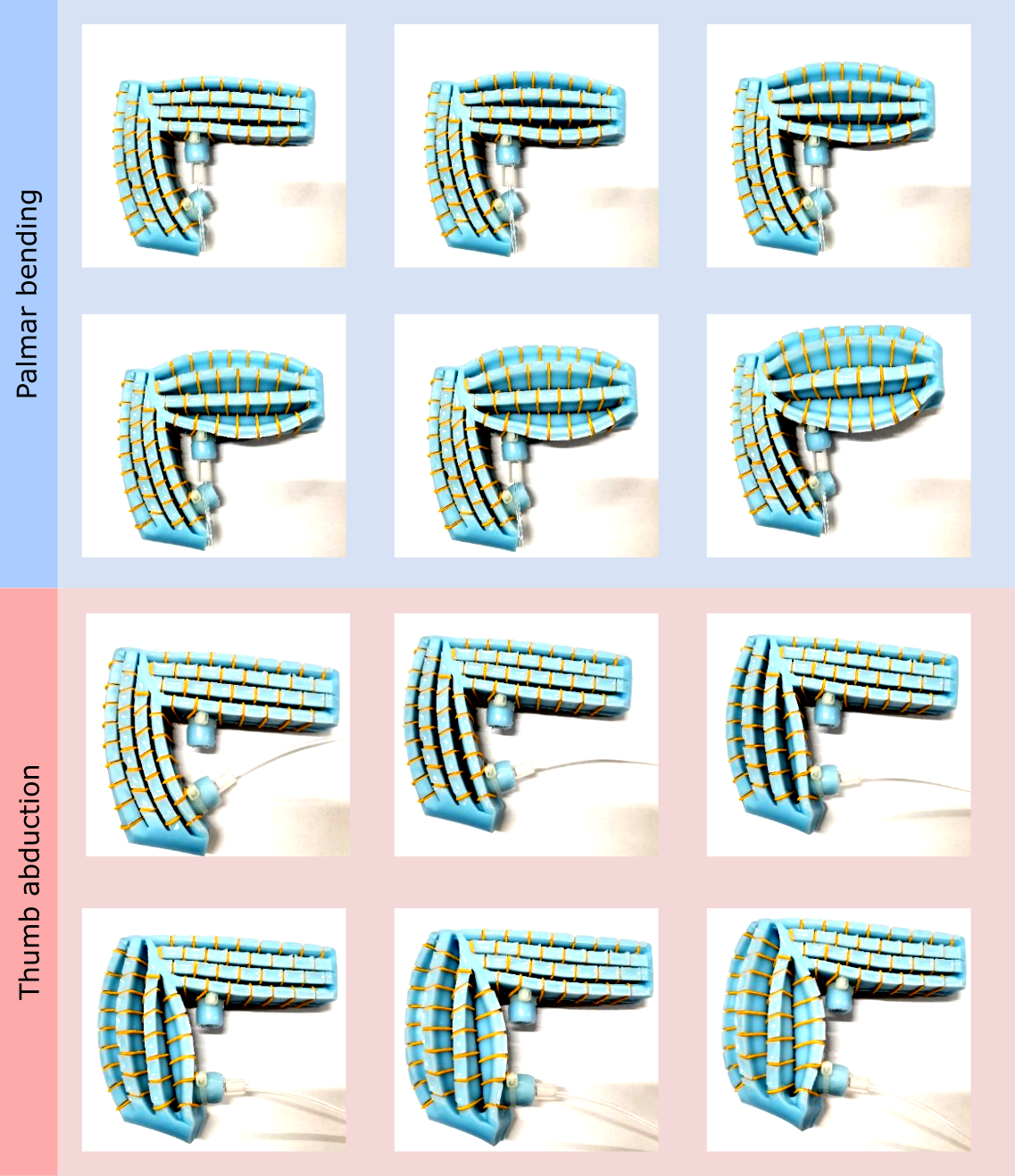}
    \caption{The test results of the soft finger}
    \label{fig:SPtest}
\end{figure}

The human-inspired soft palm used in this study was tested to analyze its performance on palmar bending and thumb abduction. As shown in Figure \ref{fig:SPtest}, the soft palm module was test and its two soft actuators were driven respectively. The inflating motions by the soft actuators will be transferred as the bending motions of the palm and thumb. And the compression force generated by the inflated chambers will be transferred as the grasping force of the soft palm.

\subsection{Water hydraulic valve control system}

\begin{figure}[htbp]
    \centering
    \includegraphics[width=1\linewidth]{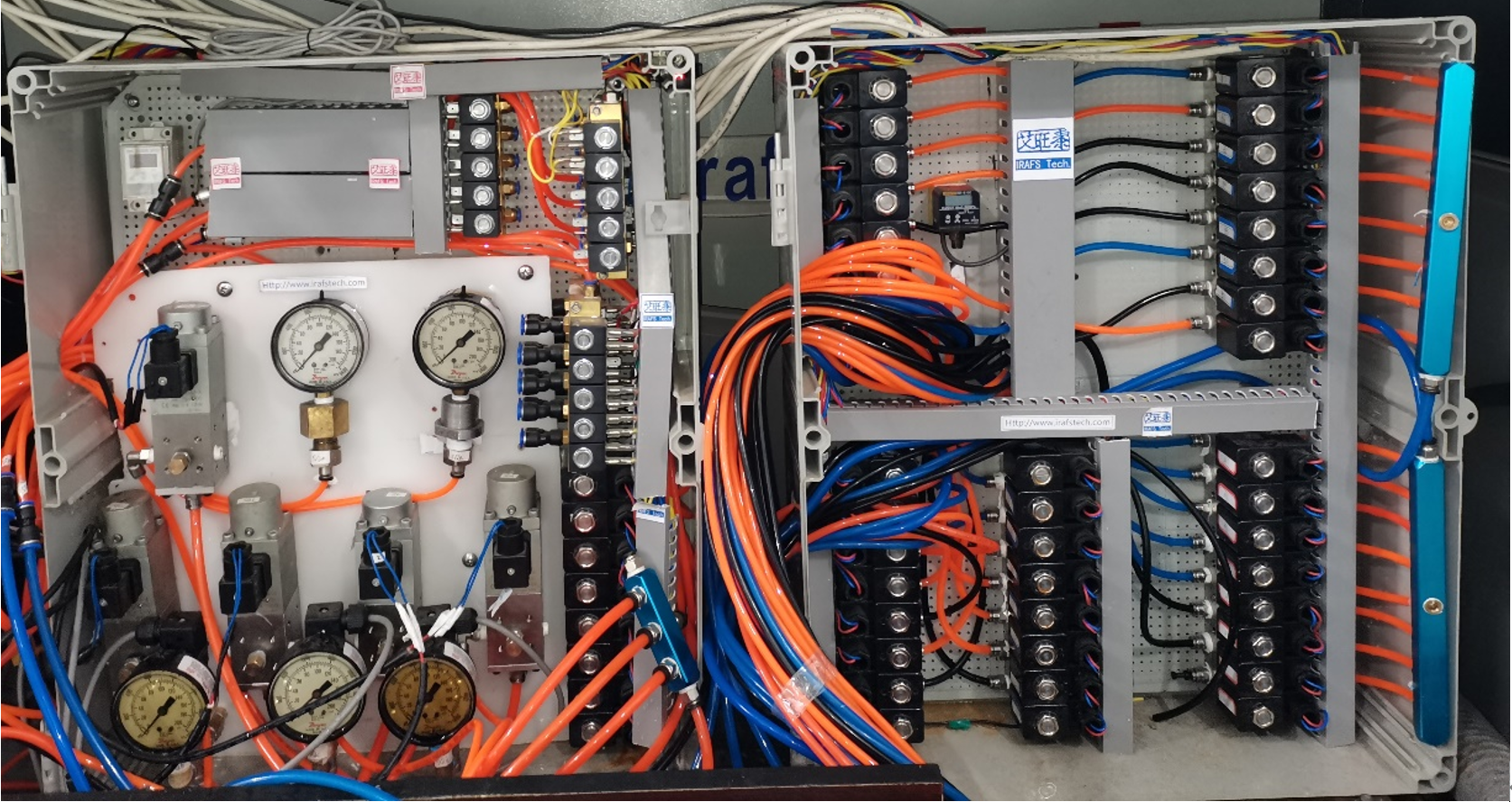}
    \caption{The water hydraulic valve control system}
    \label{fig:ValveControl}
\end{figure}
The experimental setups for the pressure regulation characteristics of the water pressure proportional pressure valve is shown in Figure \ref{fig:ValveControl}. It is mainly composed of a control cabinet, a computer, a proportional amplifier, an analog input and output module, a proportional pressure control experimental valve, and a display instrument device. The proportional pressure control experiment valve is connected to the piping system of the water hydraulic experiment bench through a pneumatic quick connector. The water hydraulic test bench can provide a pressure of 6 MPa. The inlet pressure of the test valve can be set by adjusting the working frequency of the motor in the control cabinet and the overflow valve in the piping system of the test bench. The outlet pressure of the test bench can be obtained on the display instrument. pressure.

\subsection{Experimental setups}

\begin{figure}[tbp]
    \centering
    \includegraphics[width=1\linewidth]{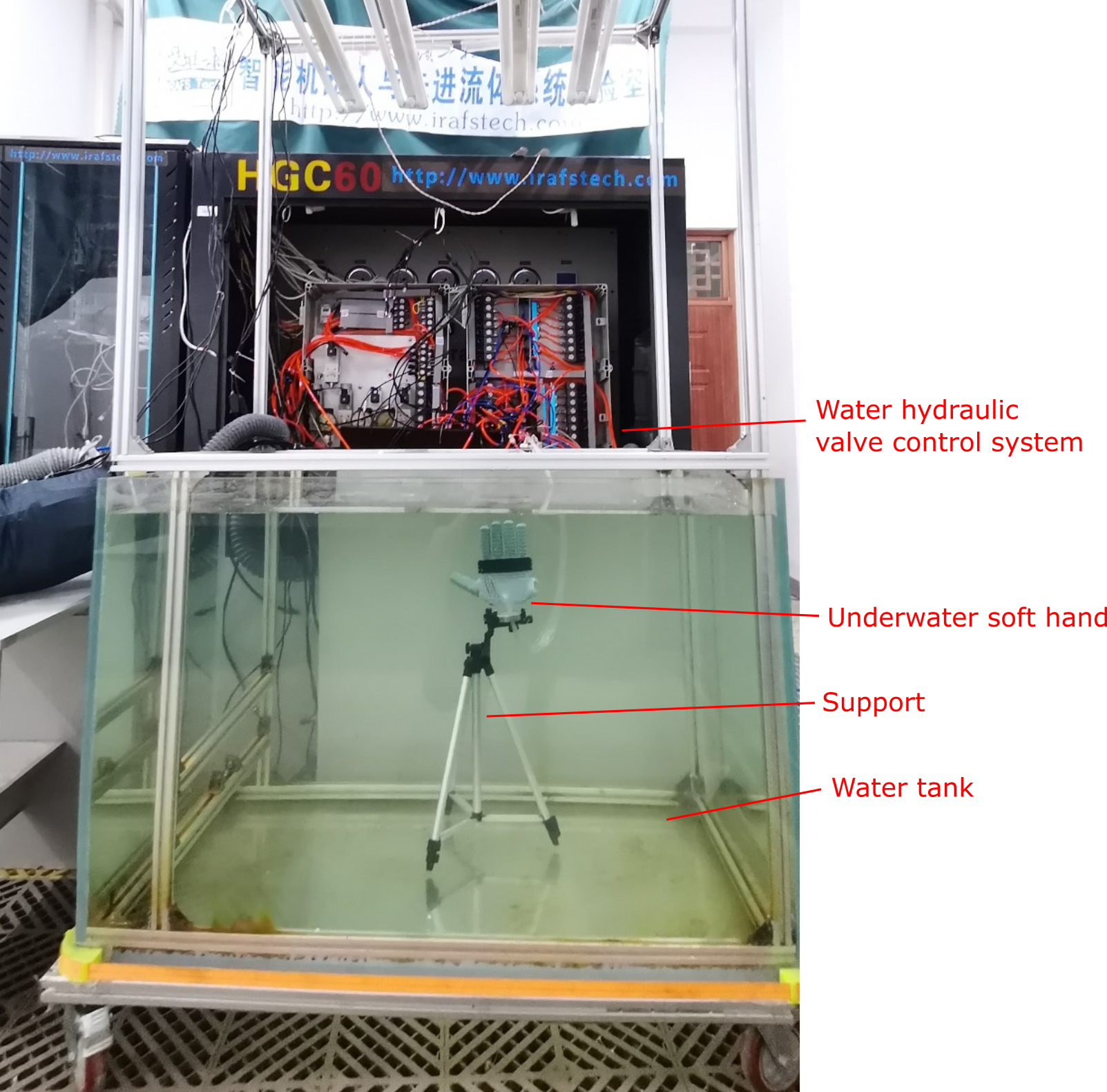}
    \caption{The experimental setups of the underwater soft robotic hand}
    \label{fig:Setups}
\end{figure}

Figure \ref{fig:Setups} shows the experimental setups of the underwater soft robotic hand.
It mainly includes software robot hand, fine hydraulic control system, water tank, bracket, water hydraulic test bench, electric control system, etc. Fill the water tank with water, even if the soft robot hand is working in an underwater environment, so as to test its effectiveness in underwater operations.

Based on the laboratory water hydraulic test system, the experimental platform of the software robot hand, the water pressure proportional pressure control valve, and the water pressure solenoid switch valve that can be used in this paper is built. The soft robot hand is fixed by a bracket and placed in an underwater environment. Each soft finger has 4 independent cavities, so 4 independent pipelines are required for pressurization control. When the software robot hand performs the grasping operation, through the precision hydraulic control system, the pressure adjustment and switch control of the independent chambers of the soft robotic hand are carried out, so that the fingers and the software palm cooperate to achieve grasping of the target object.

\section{Conclusions}
\label{sec:concl}
A new fluid-driven soft robot hand in this study uses the idea of the bionics and has the anthropomorphic form, which is oriented to the flexible grasp function. The soft robot hand is composed of a new kind of multi-freedom soft finger and soft palm, which realizes the characteristic grasping function of forehand and backhand. Combined with the fine fluid control system, the soft hand can realize flexible grasping under high pressure, so as to realize flexible grasping operation for different types of target objects in the underwater environment. According to the existing water hydraulic valve technology, a new proportional pressure control valve was designed by using the hydraulic half bridge circuit to control its pressure. The water hydraulic proportional throttle valve was selected as the adjustable fluid resistance in the C hydraulic half axle and the manual throttle valve as the fixed fluid resistance. According to the design parameters of the proportional throttle valve structure design, the proportional throttle valve and manual throttle valve were integrated into a valve to form a new proportional pressure control valve based on the C-type hydraulic half bridge. Five new proportional pressure regulating valves were made, and a fine fluid control system was formed. Finally, the soft robot hand and the fine fluid control system were connected to form the underwater soft robot hand experiment platform.

\section*{ACKNOWLEDGMENT}
This work was supported by the Natural Science Foundation of China under Grant 51875113, Natural Science
Joint Guidance Foundation of the Heilongjiang Province of China under Grant LH2019E027.

	


	%
	%
	%
	%
	%
	%
	
	
	\bibliographystyle{IEEEtran}
	\bibliography{ref}

\end{document}